\documentclass[journal]{IEEEtai}

\usepackage[colorlinks,urlcolor=blue,linkcolor=blue,citecolor=blue]{hyperref}

\usepackage{color,array}

\usepackage{graphicx}

\usepackage{amsmath,amsfonts}
\usepackage{algorithmic}
\usepackage{array}
\usepackage[caption=false,font=normalsize,labelfont=sf,textfont=sf]{subfig}
\usepackage{textcomp}
\usepackage{stfloats}
\usepackage{url}
\usepackage{verbatim}
\usepackage{graphicx}
\usepackage{hyperref}
\usepackage{multirow}

\usepackage{balance}
\begin{document}
\title{SAMScore: A Content Structural Similarity Metric for Image Translation Evaluation}
\author{Yunxiang Li, Meixu Chen, Kai Wang, Jun Ma, Alan C. Bovik, \IEEEmembership{Life Fellow, IEEE}, You Zhang
\thanks{This work was supported by the National Institutes of Health (Grant No. R01 EB034691, R01 CA240808, R01 CA258987, and R01 CA280135). Corresponding author: You Zhang (e-mail: you.zhang@utsouthwestern.edu)}   
\thanks{Yunxiang Li, Meixu Chen, Kai Wang, You Zhang are with Department of Radiation Oncology, The University of Texas Southwestern Medical Center, Dallas, TX, 75390, USA.}
\thanks{Jun Ma is with Department of Laboratory Medicine and Pathobiology, University of Toronto,
Toronto, Canada. }
\thanks{Alan C. Bovik is with Laboratory for Image and Video Engineering, The University of Texas at Austin, Austin, TX, 78712, USA }  
}

\markboth{Journal of IEEE Transactions on Artificial Intelligence, Vol. 00, No. 0, January 2025}
{Li \MakeLowercase{\textit{et al.}}: A Content Structural Similarity Metric for Image Translation Evaluation}

\maketitle

\begin{abstract}
Image translation has wide applications, such as style transfer and modality conversion, usually aiming to generate images having both high degrees of realism and faithfulness. These problems remain difficult, especially when it is important to preserve content structures. Traditional image-level similarity metrics are of limited use, since the content structures of an image are high-level, and not strongly governed by pixel-wise faithfulness to an original image. To fill this gap, we introduce SAMScore, a generic content structural similarity metric for evaluating the faithfulness of image translation models. SAMScore is based on the recent high-performance Segment Anything Model (SAM), which allows content similarity comparisons with standout accuracy. We applied SAMScore on 19 image translation tasks, and found that it is able to outperform all other competitive metrics on all tasks. We envision that SAMScore will prove to be a valuable tool that will help to drive the vibrant field of image translation, by allowing for more precise evaluations of new and evolving translation models.
\end{abstract}

\begin{IEEEImpStatement}
Image translation is the process of changing the style or domain of an image while keeping its essential content intact. It has wide and varied implications, especially in the fields of art and medical imaging. For example, in the context of art, image translation is often used for stylistic transformations to superimpose the beauty of a famous painting onto a photograph. In the field of medical imaging, it allows the conversion of magnetic resonance imaging (MRI) to computed tomography (CT) scan, which is widely used in radiation therapy. However, current metrics typically do not adequately reflect the preservation of complex content structures. Our innovative metrics are expected to set new standards in the field and facilitate the advancement of various applications. This will further enhance the efficiency gains of AI in art creation, medical imaging, and more.
\end{IEEEImpStatement}


\section{Introduction}
\IEEEPARstart{I}{mage} translation, including tasks such as style transfer and modality conversion, has been a highly active research area in computer vision and deep learning~\cite{isola2017image, liu2017unsupervised, zhu2017toward, choi2018stargan, richardson2021encoding, armanious2020medgan,tumanyan2023plug,li2023bbdm,lee2024conditional,xu2024cyclenet}. 
Image translation tasks aim to generate images in a target domain while ensuring the adequate preservation of content structural information intrinsic to the source image~\cite{pang2021image,kaji2019overview,alotaibi2020deep}. 
The diverse applications of image translation, ranging from medical image processing~\cite{armanious2020medgan} to autonomous driving~\cite{arar2020unsupervised}, have fueled significant interest in this field. Across these many domains, one of the primary challenges in achieving high-quality image translation is the preservation of content structure in the generated images (faithfulness)~\cite{zhang2018unreasonable, borji2019pros, bashkirova2022evaluation}. However, there is a lack of accepted or effective metrics for evaluating the ability of image translation to produce results that exhibit satisfactory content structure preservation. 
We believe that addressing this missing need will prove to be an essential ingredient for advancing the field of image translation.

In image translation tasks, the role of metrics can generally be divided into two categories: optimization metrics as loss functions and evaluation metrics for model performance analysis. When used as loss functions, metrics should present differentiability and stability to enable deep learning models to effectively optimize the objective function through gradient descent methods. On the other hand, metrics used for model performance analysis focus more on the interpretability of the results and their applicability to real-world scenarios. These metrics are primarily designed to quantify the similarity between generated and target images, without considering their differentiability during the model training process. Conventional image-level similarity metrics, such as the L2 norm, Peak Signal-to-Noise Ratio (PSNR), and Structural Similarity Index (SSIM)~\cite{wang2004image}, despite their ability to measure image similarity, are inadequate for this purpose. Their deficiency lies in their inability to incorporate high-level content understanding into their measurements in a manner that does not require pixel-to-pixel similarity~\cite{lahouhou2010selecting}. This flaw becomes evident when considering the measurement of the translation from a black cat to a white cat using any of the aforementioned similarity models. Likewise, the Learned Perceptual Image Patch Similarity (LPIPS) metric~\cite{zhang2018unreasonable}, which does utilize high-level information, falls short as well since it is not designed with the purpose of evaluating content structures. LPIPS measures only perceived similarities between images in the presence of common distortions, which is rather limited. FCNScore~\cite{borji2019pros, zhu2017unpaired, isola2017image}, which employs fully convolutional networks for image content segmentation and indirectly assesses the structural similarity via the resulting segmentations, has its own set of issues. These include the necessity of labels to train segmentation networks, a lack of granularity of structure evaluation, and the compounding effects of potential domain gaps between the generated images and the target images (FCNScore is trained on the target images, and the domain gap will affect the segmentation accuracy on the generated images).

Some researchers have attempted to leverage large models to construct metrics for more precise evaluations, such as BERTScore \cite{zhangbertscore} for text similarity and CLIPScore \cite{hessel2021clipscore} for evaluating text-to-image generation. The recently introduced Segment Anything Model (SAM)~\cite{kirillov2023segment, MedSAM} is a highly powerful segmentation method, exhibiting strong zero-shot generalization capabilities, with the ability to extract highly generalizable structural information. We exploit this timely advancement to develop a new image similarity metric that we call SAMScore, which leverages SAM to evaluate content structural similarities between original source images and generatively-translated versions of them. The primary distinction between SAMScore and SAM lies in their intended purposes and functionalities. SAM is designed as a general-purpose segmentation model aimed at segmenting arbitrary objects in images, excelling in zero-shot generalization and cross-domain scalability. However, SAM itself is not designed to evaluate the structural similarity or quality of generated images. In contrast, SAMScore focuses on utilizing the highly generalizable structural information extracted by SAM to quantify the content structural similarity between original and generated images. Our contributions are summarized as follows:
\begin{itemize}
    \item We propose a universal metric for assessing content shape structural similarity, which addresses the current lack of any general metric for evaluating the content faithfulness of image translation tasks.
    \item We show that SAMScore outperforms existing similarity metrics in evaluating content shape structural similarity across 19 image translation tasks, demonstrating superior effectiveness and robustness.
\end{itemize}

\begin{figure*}[]
  \centering
  \includegraphics[width=1.9\columnwidth]{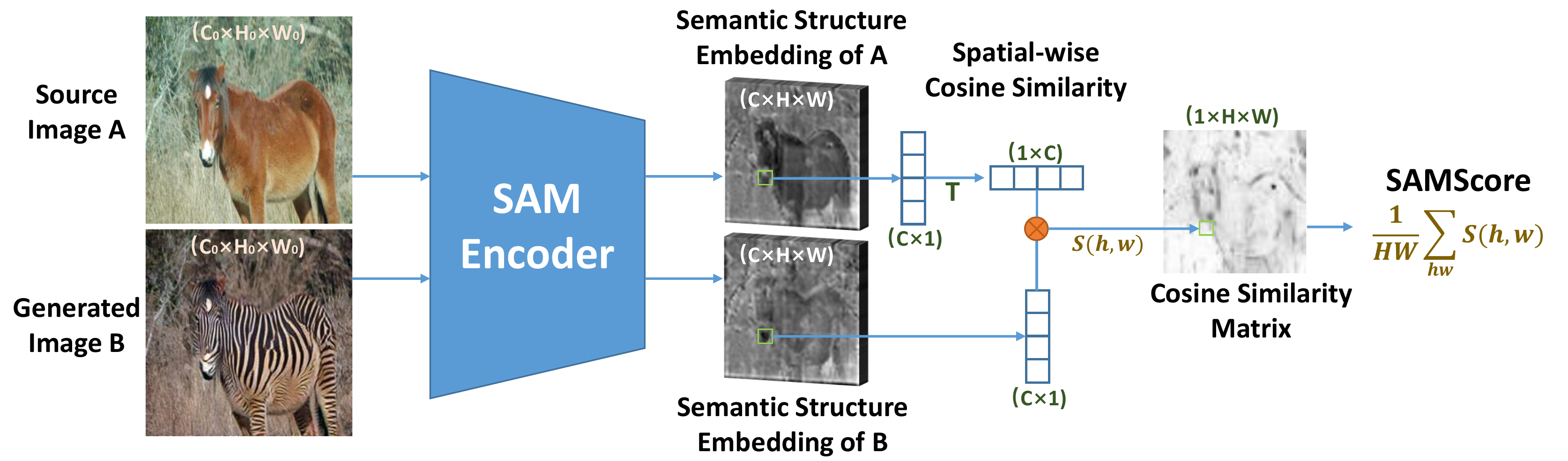}
  \caption{Overview of SAMScore. The source and the generated images are separately input to the SAM encoder to obtain embeddings of the content structures. A spatial cosine similarity score is then calculated, yielding the final SAMScore.}
  \label{fig:overview}
\end{figure*}

\section{Related Work}
\label{sec:rel_work}

The primary goal of image translation is to map and transform images between different domains. Although various techniques have been developed, their core objective remains the same, which is to achieve realistic and faithful image translations. Faithfulness here refers to the preservation of the content structure information from the source image. Previous methods for evaluating faithfulness include L2 (Euclidean distance or MSE), PSNR (Peak Signal-to-Noise Ratio), SSIM (Structural Similarity Index)~\cite{wang2004image},  LPIPS (Learned Perceptual Image Patch Similarity)~\cite{zhang2018unreasonable}, and FCNScore (content segmentation accuracy by Fully Convolutional Network)~\cite{borji2019pros, zhu2017unpaired, isola2017image}. 

\textbf{L2} is a widely used metric that compares images pixel by pixel. Despite being simple and easy to calculate, L2 has limitations that make it unsuitable for most image translation tasks. For instance, it is sensitive to image intensity range shifts, causing significant differences even if the content/structure of the images remains similar.

\textbf{PSNR} is a common metric often used to assess image compression. It is equivalent to L2 and also falls short in assessing content structural similarity.

\textbf{SSIM} captures perceptual similarity more accurately than L2/PSNR by accounting for perceptual masking and low-level local textures~\cite{wang2004image}. However, SSIM lacks the ability to capture high-level structural information for comparison.

\textbf{LPIPS} is a deep learning-based metric that measures the distance or difference between image patches, by computing the similarity between the activations of images output by a neural network~\cite{zhang2018unreasonable}. While LPIPS accurately predicts perceptual similarity between images, it does not capture content structural similarity.

\textbf{FCNScore} is an image translation evaluation metric based on the segmentation accuracy~\cite{borji2019pros, zhu2017unpaired, isola2017image}. This metric requires training a fully connected segmentation network on target domain images for a specific segmentation task. It also requires segmentation labels to be available on the source domain images. Content structure similarity is measured by the accuracy and IoU between the segmentation labels of the source images and the segmented results by the network on the generated images. However, FCNScore incurs a high cost of targeted training, and its measurement is limited by the granularity of the segmentation labels. The potential domain gap between the target domain and the generated image domain may also affect the accuracy of the trained segmentation network (on the target domain images) and distort the FCNScore.

Evaluating the content structural faithfulness of image translation has been a challenging task, with previous studies adopting different approaches. Some, such as GcGAN~\cite{zhang2019gcgan} and CUT~\cite{park2020contrastive}, employ the FCNScore metric to assess content structural similarity. MUNIT~\cite{huang2018multimodal} and StarGANv2~\cite{choi2020stargan} use human evaluations obtained via the Amazon Mechanical Turk (AMT)~\cite{paolacci2010running}, but this introduces significant subjectivity and scalability challenges. CycleGAN~\cite{zhu2017unpaired} and pix2pix~\cite{isola2017image} combine FCNScore and AMT to leverage the strengths of both metrics, but this method still has intrinsic limitations from both metrics. In recent diffusion model-based methods, the used datasets usually lack available segmentation labels, hence L2, PSNR, SSIM, and LPIPS metrics are used instead of FCNScore. BBDM~\cite{li2022vqbb} uses LPIPS for evaluation; SDEdit~\cite{meng2021sdedit} applies L2 and LPIPS; CycleDiffusion~\cite{wu2022unifying} employs PSNR and SSIM; and EGSDE~\cite{zhao2022egsde} uses L2 along with PSNR and SSIM. Although these models have successfully measured differences to some extent, they do not specifically evaluate content structural similarity.

\section{SAMScore}
\label{headings}

To address the limitations of the existing similarity metrics described in the previous section, we introduce SAMScore, which uses the encoder portion of the foundational Segment Anything Model (SAM) to obtain rich content structural embeddings of both source and generated images for comparison. SAMScore measures content structural similarity by calculating the cosine similarity between the two resulting embeddings, as shown in Fig. \ref{fig:overview}.

\begin{figure*}[hb]
  \centering
  \includegraphics[width=.99\textwidth]{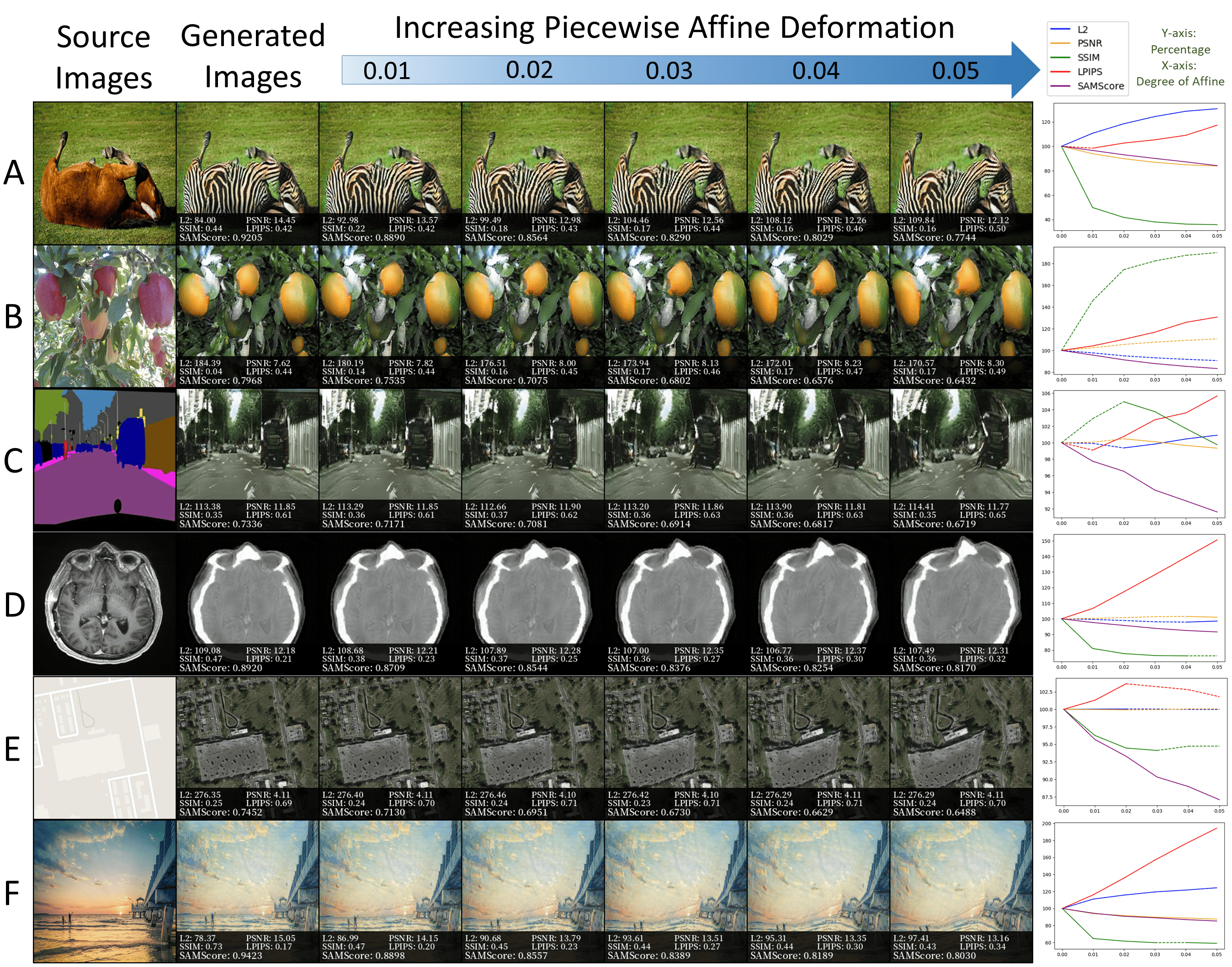}
  \caption{Similarity scores between the original images and CycleGAN-generated images with varying degrees of deformation. (A) horse to zebra, (B) orange to apple, (C) cityscapes (label to photo), (D) head (MR to CT), (E) photo to Ukiyoe, and (F) photo to Monet. Column 2 shows the results without additional distortion. Given the differing scales of the metrics, we converted each metric to a percentage change relative to the initial value (results without any perturbation) before generating the line plots. We used dashed lines to represent trends that deviate from the ideal direction of change. }
  \label{fig:cycle_deformation}
\end{figure*}

\subsection{SAM Encoder}
We first map the source and the translated (generated) images into high-level image embedding spaces that are rich in content structural information. We deploy the SAM Encoder $\mathcal{E}(\cdot)$ to extract content embeddings of both the source image and the translated image, denoted as $X_E$ and $Y_E$, respectively. The large and extensive prompt-based segmentation data used to train the SAM encoder have equipped it with the capability to extract general images’ high-level structures. Instead of low-level local color or texture information, the SAM encoder focuses on extracting the content structural information for the segmentation needs.
Given a source image $X_0\in \mathbb{R}^{C_0 \times H_0 \times W_0}$ and a translated image $Y_0 \in \mathbb{R}^{C_0\times H_0 \times W_0}$, let:

\begin{equation}
\label{eq1}
    X_E=\mathcal{E}(X_0),~ Y_E=\mathcal{E}(Y_0),
\end{equation}

where $X_E, Y_E \in \mathbb{R}^{C \times H \times W}$ are the respective embeddings, and $C$, $H$, and $W$ are the number of channels, the height, and the width of the output embeddings of the SAM encoder, respectively. In the standard SAM encoder, $C$ is 256, and $H$ and $W$ are equal to 1/16 of the input image’s height $H_0$ and width $W_0$, respectively.

\subsection{Similarity Metrics} 
After obtaining the SAM encoder embedding of the source and the translated images, we measure their similarity by the cosine similarity, which is a simple but effective target to compare the similarity of two vectors. Cosine similarity measures the cosine of the angle between two vectors, taking into account the direction of the vectors rather than their size. This provides robustness when comparing tensor features in the presence of small amounts of noise or variations of magnitudes, especially for high-dimensional data. Since content structural information is spatial, we calculate the similarity of the two embeddings at the spatial level. The cosine similarity between the two vectors in Equation \ref{eq1} at spatial position $(h,w)$ is:

\begin{equation}
S(X_E^{hw},Y_E^{hw}) = \frac{(X_E^{hw})^\top Y_E^{hw}}{||X_E^{hw}||_2~||Y_E^{hw}||_2}.
\end{equation}

The overall  SAMScore between the source image and the generated image is then:

\begin{equation}
\begin{aligned}
SAMScore(X_E,Y_E) &=  \frac{1}{HW}\sum_{h,w}S(X_E^{hw},Y_E^{hw}) \\
&= \frac{1}{HW}\sum_{h,w} \frac{(\mathcal{E}^{hw}(X_0))^\top \mathcal{E}^{hw}(Y_0) }{||\mathcal{E}^{hw}(X_0)||_2~||\mathcal{E}^{hw}(Y_0)||_2}.
\end{aligned}
\end{equation}

\begin{table*}[h]
\centering
\renewcommand\arraystretch{1.0}
\setlength{\tabcolsep}{5mm}
\caption{Absolute correlation coefficients of image similarity metrics on a variety of image translation tasks, where the images have been corrupted by varying amounts of affine geometric distortions. Since deformation changes image content structural information, a higher correlation is better as it shows a metric can better capture the content structural differences between the source and translated images. Best value is shown in bold.}
\begin{tabular}{c|ccccc}
\hline
Task                    & L2      & PSNR    & SSIM    & LPIPS  & SAMScore \\ \hline
apple to orange            & 0.7817 & 0.7810 & 0.8223 & 0.8871 & \textbf{0.9116} \\
orange to apple             & 0.7967 & 0.7958 & 0.8390 & 0.9134 & \textbf{0.9315} \\
cityscapes (label to photo) & 0.4577 & 0.4570 & 0.4914 & 0.7801 & \textbf{0.9136} \\
cityscapes (photo to label) & 0.4635 & 0.4633 & 0.4824 & 0.6150 & \textbf{0.7892} \\
facades (label to photo)   & 0.7460 & 0.7457 & 0.6943 & 0.7740 & \textbf{0.9549} \\
facades (photo to label)   & 0.7739 & 0.7734 & 0.7087 & 0.7095 & \textbf{0.9479} \\
head (MR to CT)               & 0.5279 & 0.5278 & 0.7134 & 0.7655 & \textbf{0.7734} \\
head (CT to MR)          & 0.7281 & 0.7304 & 0.7798 & 0.8208 & \textbf{0.8553} \\
horse to zebra             & 0.9276 & 0.9183 & 0.8190 & 0.9169 & \textbf{0.9497} \\
zebra to horse               & 0.9337 & 0.9235 & 0.8334 & 0.9446 & \textbf{0.9564} \\
Monet to photo           & 0.8548 & 0.8377 & 0.7701 & 0.9395 & \textbf{0.9444} \\
photo to Monet           & 0.9088 & 0.8992 & 0.7895 & 0.9386 & \textbf{0.9487} \\
aerial photograph to map                & 0.7275 & 0.7275 & 0.7827 & 0.7997 & \textbf{0.9326} \\
map to aerial photograph            & 0.5481 & 0.5483 & 0.7321 & 0.7984 & \textbf{0.9317} \\
photo to Cezanne            & 0.9240 & 0.9155 & 0.8178 & 0.9161 & \textbf{0.9338} \\
photo to Ukiyoe          & 0.8963 & 0.8917 & 0.7798 & 0.9148 & \textbf{0.9261} \\
photo to Vangogh          & 0.9109 & 0.9009 & 0.7745 & 0.9137 & \textbf{0.9273} \\
Yosemite (summer to winter) & 0.9183 & 0.9201 & 0.6889 & 0.9372 & \textbf{0.9545} \\
Yosemite (winter to summer) & 0.9205 & 0.9224 & 0.7297 & 0.9399 & \textbf{0.9488} \\ 
\hline
\end{tabular}
\label{corr_deformation}
\end{table*}

\begin{figure*}[h]
  \centering
  \includegraphics[width=.99\textwidth]{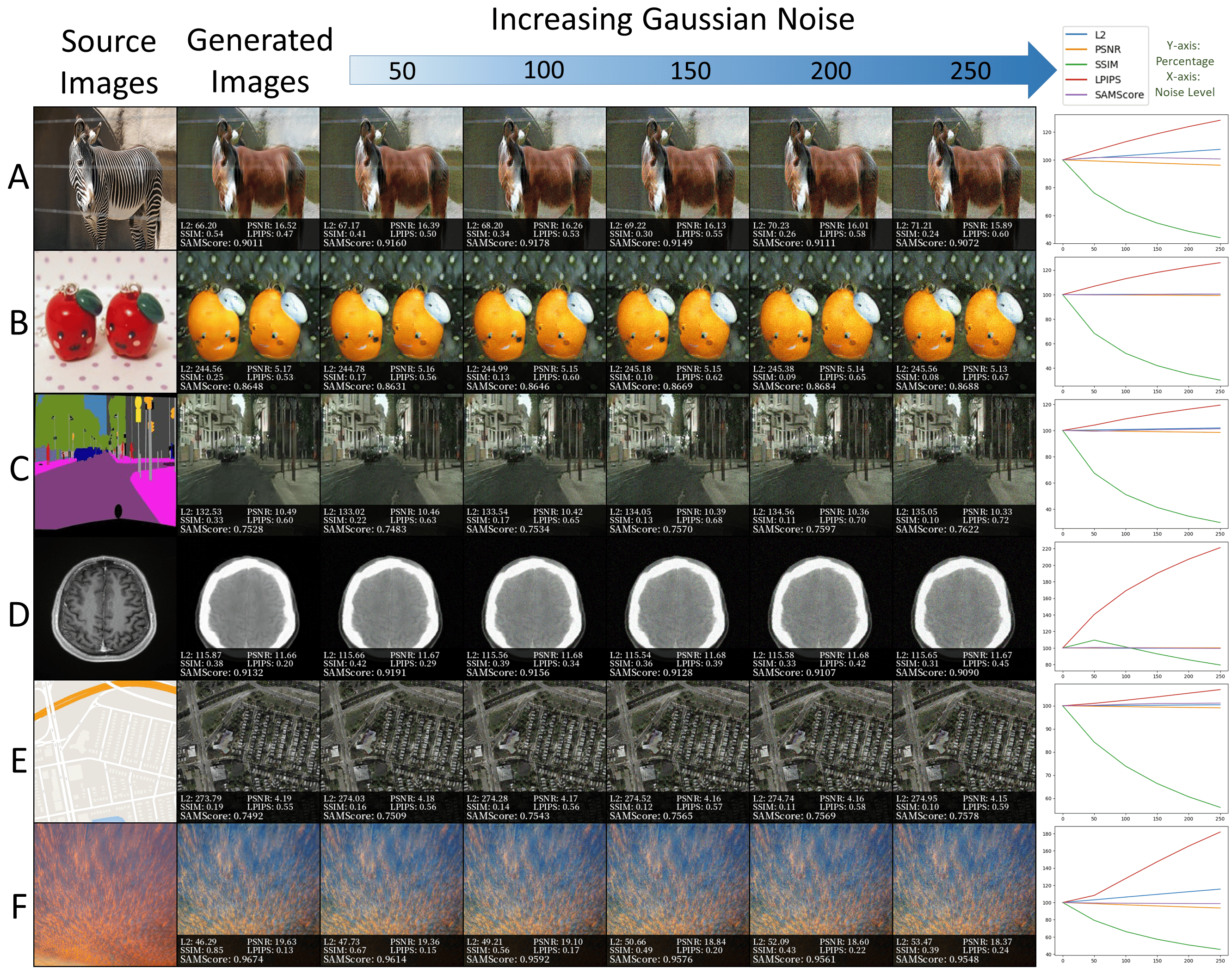}
  \caption{Similarity scores between the original images and CycleGAN-generated images with varying degrees of Gaussian noise corruption. (A) horse to zebra, (B) orange to apple, (C) cityscapes (label to photo), (D) head (MR to CT), (E) photo to Ukiyoe, and (F) photo to Monet. Column 2 shows results without additional distortion. Given the differing scales of the metrics, we converted each metric to a percentage change relative to the initial value (results without any perturbation) before generating the line plots.}
  \label{fig:cycle_gaussian}
\end{figure*}

\begin{table*}[h]
\centering
\renewcommand\arraystretch{1.0}
\setlength{\tabcolsep}{5mm}
\caption{Absolute correlation coefficients of image similarity metrics on a variety of image translation tasks, where the images have been corrupted by varying degrees of Gaussian noise. Since Gaussian noise does not significantly change image content structural information, unless it is very heavy, a lower correlation is better as it shows the metric is less susceptible to random noise-induced intensity variations between the source and the translated images. Best value is shown in bold.}
\begin{tabular}{c|ccccc}
\hline
Task                    & L2      & PSNR    & SSIM    & LPIPS  & SAMScore \\ \hline
apple to orange           & 0.8180 & 0.8179 & 0.9200 & 0.9435 & \textbf{0.6538} \\
orange to apple            & 0.8511 & 0.8511 & 0.9220 & 0.9506 & \textbf{0.6953} \\
cityscapes (label to photo) & 0.9950 & 0.9948 & 0.9269 & 0.9925 & \textbf{0.3881} \\
cityscapes (photo to label)& 0.9845 & 0.9847 & 0.9255 & 0.9855 & \textbf{0.4935} \\
facades (label to photo)  & 0.9392 & 0.9392 & 0.9444 & 0.9300 & \textbf{0.7025} \\
facades (photo to label)   & 0.9693 & 0.9698 & 0.9523 & \textbf{0.7487} & 0.7603 \\
head (MR to CT)            & 0.7873 & \textbf{0.7852} & 0.9227 & 0.9777 & 0.8248 \\
head (CT to MR)            & 0.9737 & 0.9731 & 0.9133 & 0.9878 & \textbf{0.5703} \\
horse to zebra             & 0.9975 & 0.9963 & 0.9586 & 0.8996 & \textbf{0.7928} \\
zebra to horse             & 0.9985 & 0.9980 & 0.9638 & 0.9476 & \textbf{0.7876} \\
Monet to photo            & 0.9879 & 0.9853 & 0.9609 & \textbf{0.7952} & 0.8324 \\
photo to Monet             & 0.9902 & 0.9903 & 0.9596 & 0.9669 & \textbf{0.7711} \\
aerial photograph to map                 & 0.8792 & 0.8795 & 0.9583 & 0.9814 & \textbf{0.8697} \\
map to aerial photograph                  & 0.8907 & 0.8904 & 0.9602 & 0.7700 & \textbf{0.5797} \\
photo to Cezanne         & 0.9983 & 0.9976 & 0.9575 & 0.8436 & \textbf{0.7012} \\
photo to Ukiyoe              & 0.9851 & 0.9853 & 0.9595 & 0.8385 & \textbf{0.6482} \\
photo to Vangogh        & 0.9974 & 0.9973 & 0.9722 & 0.8345 & \textbf{0.6318} \\
Yosemite (summer to winter) & 0.9985 & 0.9982 & 0.9364 & 0.9317 & \textbf{0.8234} \\
Yosemite (winter to summer) & 0.9975 & 0.9969 & 0.9405 & 0.9538 & \textbf{0.8333} \\ \hline
\end{tabular}
\label{corr_Gaussian}
\end{table*}

\begin{table*}[h]
\centering
\renewcommand\arraystretch{1.1}
\setlength{\tabcolsep}{5mm}
\caption{Absolute correlation coefficients of image similarity, measured by FCNScore and SAMScore on the cityscapes (label to photo) task, where the translated images have been distorted by varying degrees of affine geometric deformations and Gaussian Noises. Best value is shown in bold.}
\begin{tabular}{c|ccccc}
\hline
Image Transformation   & per-pixel ACC     & class IoU   & SAMScore        \\ \hline
Deformation & 0.6693 & 0.8353   &  \textbf{0.9136}  \\ 
Gaussian Noise & 0.7119 & 0.7980 & \textbf{0.3881}\\ 
 \hline
\end{tabular}
\label{corr_FCNScore}
\end{table*}

\begin{figure*}[h]
  \centering
  \includegraphics[width=.99\textwidth]{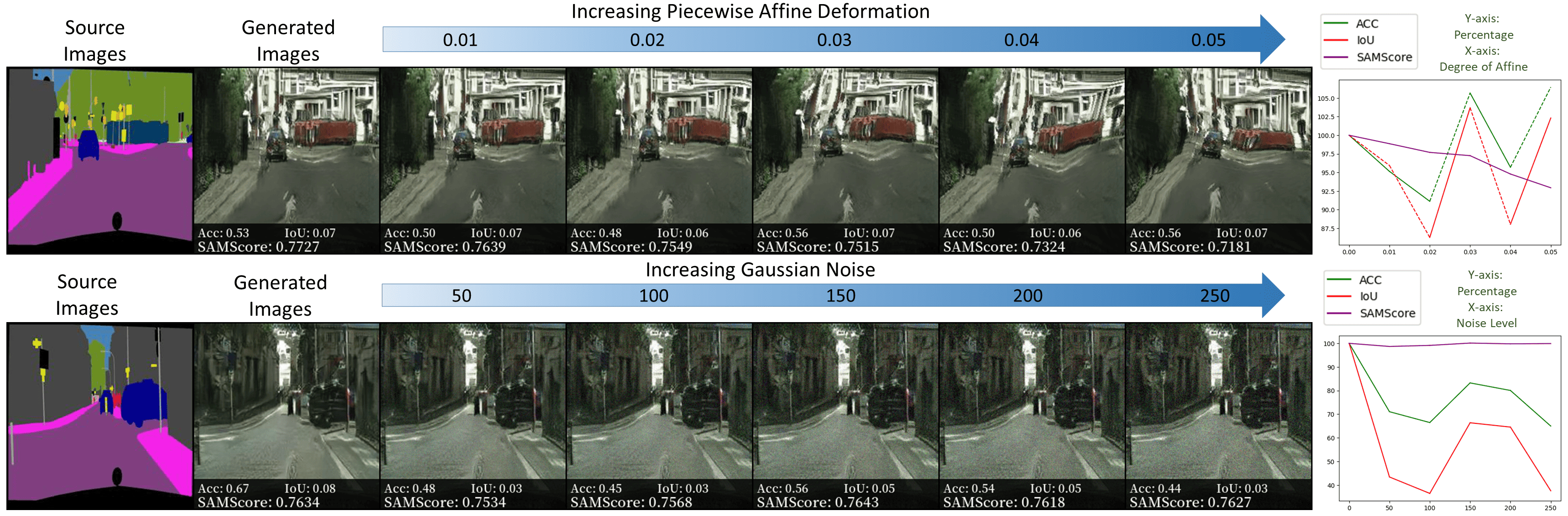}
  \caption{FCNScore and SAMScore: Piecewise affine deformations and Gaussian noises were applied to the generated images, then segmentation was performed by a DeepLabV3Plus that had been trained on target real images. The content shape structural similarity of the generated images was then evaluated by applying accuracy and IoU metrics. Column 2 shows results without added distortion. Given the differing scales of the metrics, we converted each metric to a percentage change relative to the initial value (results without any perturbation) before generating the line plots. We used dashed lines to represent trends that deviate from the ideal direction of change.}
  \label{fig:FCNScore}
\end{figure*}

\section{Experimental Setup}

We evaluated the performance of SAMScore on 19 image translation tasks across 8 datasets. The medical image dataset is from the MICCAI 2020 challenge on Anatomical Brain Barrier to Cancer Spread\footnote{\url{https://abcs.mgh.harvard.edu/index.php}}~\cite{shusharina2021segmentation}, while the other datasets were obtained from CycleGAN's official public data repository\footnote{\url{https://people.eecs.berkeley.edu/~taesung_park/CycleGAN/datasets/}}~\cite{zhu2017unpaired}. The following is the list of image translation tasks considered in this paper: (1) apple to orange,   
(2) orange to apple,        
(3) cityscapes (label to photo), 
(4) cityscapes (photo to label),
(5) facades (label to photo), 
(6) facades (photo to label),  
(7) head (MR to CT),         
(8) head (CT to MR),        
(9) horse to zebra,         
(10) zebra to horse,      
(11) Monet to photo,         
(12) photo to Monet,        
(13) aerial photograph  to map,            
(14) map to aerial photograph,           
(15) photo to Cezanne,      
(16) photo to Ukiyoe,       
(17) photo to Vangogh,     
(18) Yosemite (summer to winter), and 
(19) Yosemite (winter to summer). 

\subsection{Dataset sources and details}
\textbf{Cityscapes dataset}: The Cityscapes dataset~\cite{cordts2016cityscapes} includes 2975 pairs of training images and 500 pairs of validation images. We used the validation set from Cityscapes for testing. 

\textbf{Maps and aerial photograph dataset}: The dataset is from Google Maps~\cite{isola2017image}, and includes 1096 pairs of training images and 1098 pairs of testing images. Images were sampled from in and around New York City.

\textbf{Architectural facades dataset}: The dataset is from the CMP Facade Database~\cite{tylevcek2013spatial}, including 400 pairs of training images and 106 pairs of testing images.

\textbf{Horse and Zebra dataset}: The dataset is from the ImageNet~\cite{deng2009imagenet}, including 939 images of the horse and 1177 images of the zebra for training. As for the testing set, there are 120 images of the horse and 140 images of the zebra.

\textbf{Apple and Orange dataset}: The dataset is from the ImageNet~\cite{deng2009imagenet}, including 996 apple and 1020 orange images for training, as well as 266 apple images and 248 orange images for testing.

\textbf{Yosemite (summer and winter) dataset}: The dataset is from Flickr~\cite{liu2017unsupervised}, and the training set comprises 1273 summer images and 854 winter images, and the testing set has 309 summer images and 238 winter images.

\textbf{Photo and Art dataset}: The art images are from Wikiart~\cite{liu2017unsupervised}.
The training set size of each class is 1074 (Monet), 584 (Cezanne), 401 (Vangogh), 1433 (Ukiyoe), and 6853 (Photo). The testing set size of each class is 121 (Monet), 58 (Cezanne), 400 (Vangogh), 263 (Ukiyoe), and 751 (Photo).

\textbf{Head CT and MR dataset}: The dataset is from the MICCAI 2020 challenge: Anatomical Brain Barrier to Cancer Spread\footnote{\url{https://abcs.mgh.harvard.edu/index.php}}~\cite{shusharina2021segmentation}. MR is T1-weighted, and the training set contains 6390 CT and 6390 MR 2D slice images. The testing set has 1000 paired CT and MR 2D slice images.

\textbf{SynthRAD2023 Dataset}: In our study, we used the SynthRAD2023 dataset\footnote{\url{https://synthrad2023.grand-challenge.org/}}~\cite{thummerer2023synthrad2023}, which includes 180 pairs of 3D pelvic scans (MR and CT). For our experimental setup, we randomly selected 10 MR-CT pairs as the test set, with the remaining pairs used for training. Each 3D image was processed into 2D slices, and all slices were resized to a resolution of 256 × 256 pixels. For MR images, intensity clipping was applied to remove the bottom 0.5\% of intensity values. For CT images, Hounsfield Unit (HU) values were clipped to the range [-1000, 1000]. All images were then rescaled to [-1, 1].

\subsection{Evaluation Method}
To assess the efficacy of SAMScore for measuring content shape structural similarity between source and generated images in image translation tasks, we conducted experiments introducing two types of distortions into the generated images: geometric deformations obtained via piecewise affine transformations~\cite{pitiot2003piecewise}, and additive Gaussian noises~\cite{qin2010multivariate}. We varied the degree of each distortion and computed the absolute Pearson Correlation Coefficient between the resulting SAMScore and the level of distortion, as shown in Tables \ref{corr_deformation} and \ref{corr_Gaussian}. This allowed us to evaluate the sensitivity of SAMScore against different types of distortions, where a high correlation value indicates a higher sensitivity to such types of distortion. Tables \ref{corr_deformation} and \ref{corr_Gaussian} also contain the correlation values obtained using L2, PSNR, SSIM, and LPIPS. The correlation values of FCNScore are reported in Table \ref{corr_FCNScore}. Absolute correlation values are reported since some metrics produce higher scores with increased similarity, and others produce lower scores. In Table \ref{corr_deformation}, a correlation coefficient closer to one implies higher sensitivity to the affine distortion, which is desirable. In Table \ref{corr_Gaussian}, smaller correlations indicate lower sensitivity to noise, which is also desirable. In the case of the head CT and MR data, the test data has ground-truth images, so we additionally computed the L2 between the generated images and the ground-truth images (Table \ref{head_mr2ct}).

\subsection{Implementation Details}

We conducted image translation tasks using the publicly available, originally reported CycleGAN structure and weights, with the exception of tasks 7 and 8 which utilized medical imaging data that were not included in the original CycleGAN experiments. For tasks 7 and 8, we re-trained the CycleGAN using its default parameters and also trained recent diffusion-based translation models. The diffusion model was trained following the standard setup in ~\cite{IDDPM}, with a linear noise schedule ($\beta \in [1e-4, 0.02]$) and 1000 diffusion steps. We used the Adam optimizer ~\cite{kingma2014adam} with a batch size of 16 and a learning rate of $10^{-4}$. An Exponential Moving Average (EMA) with a rate of 0.9999 was implemented to smooth the model parameters. We resized all images to a resolution of 256×256 pixels and added the piecewise affine deformations and Gaussian noise to the generated images using the widely used Albumentations data augmentation library~\cite{buslaev2020albumentations}, which enabled us to evaluate the sensitivity and robustness of the compared image similarity assessment metrics. To generate the piecewise affine deformations, each point on the regular grid was perturbed by a random amount drawn from a normal distribution scaled to the range of [0.01, 0.05], and the number of rows and columns of the regular grid was set to 4.  The zero-mean additive Gaussian noise was drawn from a normal distribution scaled to the range of [50, 250]. To compute the SAMScore, we used the originally published weights of the SAM (ViT-L) image encoders, following the default SAM settings. Each image input to SAM was re-sampled to 1024×1024, while each embedded output from the SAM encoder was 64×64 with 256 channels. To calculate the FCNScore, we employed the DeepLabV3Plus-MobileNet segmentation model with pre-trained weights from a widely used open-source repository\footnote{\url{https://github.com/VainF/DeepLabV3Plus-Pytorch}}. Further implementation details regarding the other models utilized in this study are provided:

SDEdit\footnote{\url{https://github.com/ermongroup/SDEdit}}: SDEdit incorporates noise into the original source image through the forward diffusion, and then gradually denoises the adulterated image with a diffusion model trained on the target domain during the reverse diffusion process. When performing inference, we add noise to the source image for 500 steps.

EGSDE\footnote{\url{https://github.com/ML-GSAI/EGSDE}}: EGSDE utilizes an energy function, pre-trained on both the source and target domains, to direct the inference in image translation. The energy function is decomposed into two terms: one encourages the transferred image to drop domain-specific features to obtain realism, and the other retains domain-independent features to maintain faithfulness. When performing inference, we also add noise to the source image for 500 steps.

CycleDiffusion\footnote{\url{https://github.com/ChenWu98/cycle-diffusion}}: This method proposes the unification of potential spaces of diffusion models, which is utilized for cycle diffusion. For CycleDiffusion, a DDIM\footnote{\url{https://github.com/ermongroup/ddim}} sampler with 100 steps is employed.

Swin Transformer\footnote{\url{https://github.com/microsoft/Swin-Transformer}}: The Swin Transformer \cite{liu2021swin} is a hierarchical vision transformer model that captures both local and global features through patch-based processing and sliding windows. In this study, we modified the Swin Transformer by adding upsampling modules in the final layers to ensure the output resolution matches the input.

\section{Results and Discussion}

\subsection{Performance in the Presence of Piecewise Affine Deformations}

We studied the relationship between each similarity metric against spatial deformations. The obtained Pearson Correlation Coefficients are shown in Table \ref{corr_deformation}. Among all the compared similarity metrics, SAMScore consistently exhibited the highest correlation coefficient with the degree of geometric (piecewise affine) deformation, highlighting its exceptional sensitivity to structural deformations and its efficacy in quantifying content structural similarity. Unlike L2, PSNR, and SSIM, which directly compute pixel-level or patch-level similarities between source and translated images, and are hence greatly influenced by low-level features such as color and texture, the SAM encoder used in SAMScore was trained on large segmentation datasets to produce high-dimensional embeddings containing mainly image content structural information. LPIPS was not trained to interpret content structures, causing it to be less effective or robust.Corresponding results for some of the tasks were visualized in Fig. \ref{fig:cycle_deformation} to qualitatively support our findings. Considering the inherent directionality of each metric—e.g., lower values for L2 and LPIPS indicate better performance, while higher values for PSNR, SSIM, and SAMScore indicate better performance, we used dashed lines to represent trends that deviate from the ideal direction of change to remove the directionality ambiguity. As the degree of piecewise affine deformation was increased, structural differences between the generated and source images became more pronounced. Unlike the other similarity metrics, SAMScore steadily decreased with increasing structural deformation, showing its sensitivity to content structural changes. More detailed results can be found in Appendix C.

\subsection{Performance in the Presence of Gaussian Noise}
We distorted the translated images with varying levels of Gaussian noises, which have a negligible impact on the structural information of images when kept at a reasonable level. We measured the similarity scores of the generated images for each translation task using different metrics and calculated the corresponding Pearson correlation coefficients against the noise level. Table \ref{corr_Gaussian} summarizes the results. Our findings show that SAMScore is highly robust against Gaussian noise, as demonstrated by its consistently low correlation coefficients across most of the tasks. By contrast, traditional image similarity metrics including L2, PSNR, and SSIM exhibited strong correlations with the noise level. To supplement our quantitative results, we also provide visual comparisons of six image translation examples in Fig. \ref{fig:cycle_gaussian}. The SAMScore exhibits no apparent correlation with the degree of Gaussian noises.

\begin{figure*}[!htb]
  \centering
  \includegraphics[width=.99\textwidth]{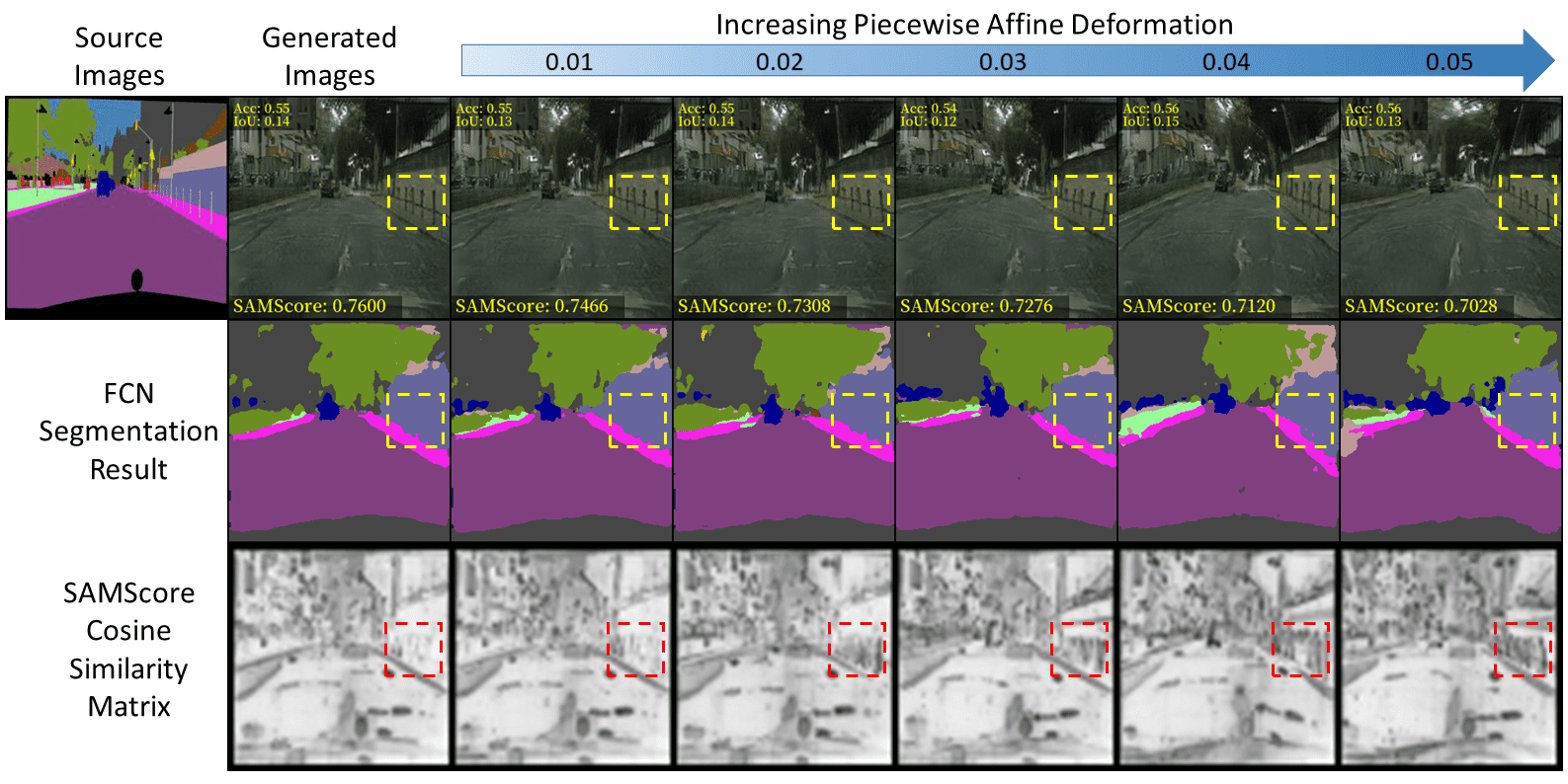}
  \caption{FCNScore and SAMScore: Piecewise affine deformations were applied to the generated images of the cityscapes (label to photo) task, and then segmentation was performed by a FCN that had been trained on `true' target images. The content shape structural similarity of the generated images was then evaluated by applying accuracy and IoU metrics. Column 2 shows results without added distortion. Row 2 shows the segmentation results of FCN, and row 3 shows the cosine similarity matrix of SAMScore.}
  \label{fig:appendix_fcn1}
\end{figure*}

\begin{figure*}[!htb]
  \centering
  \includegraphics[width=.99\textwidth]{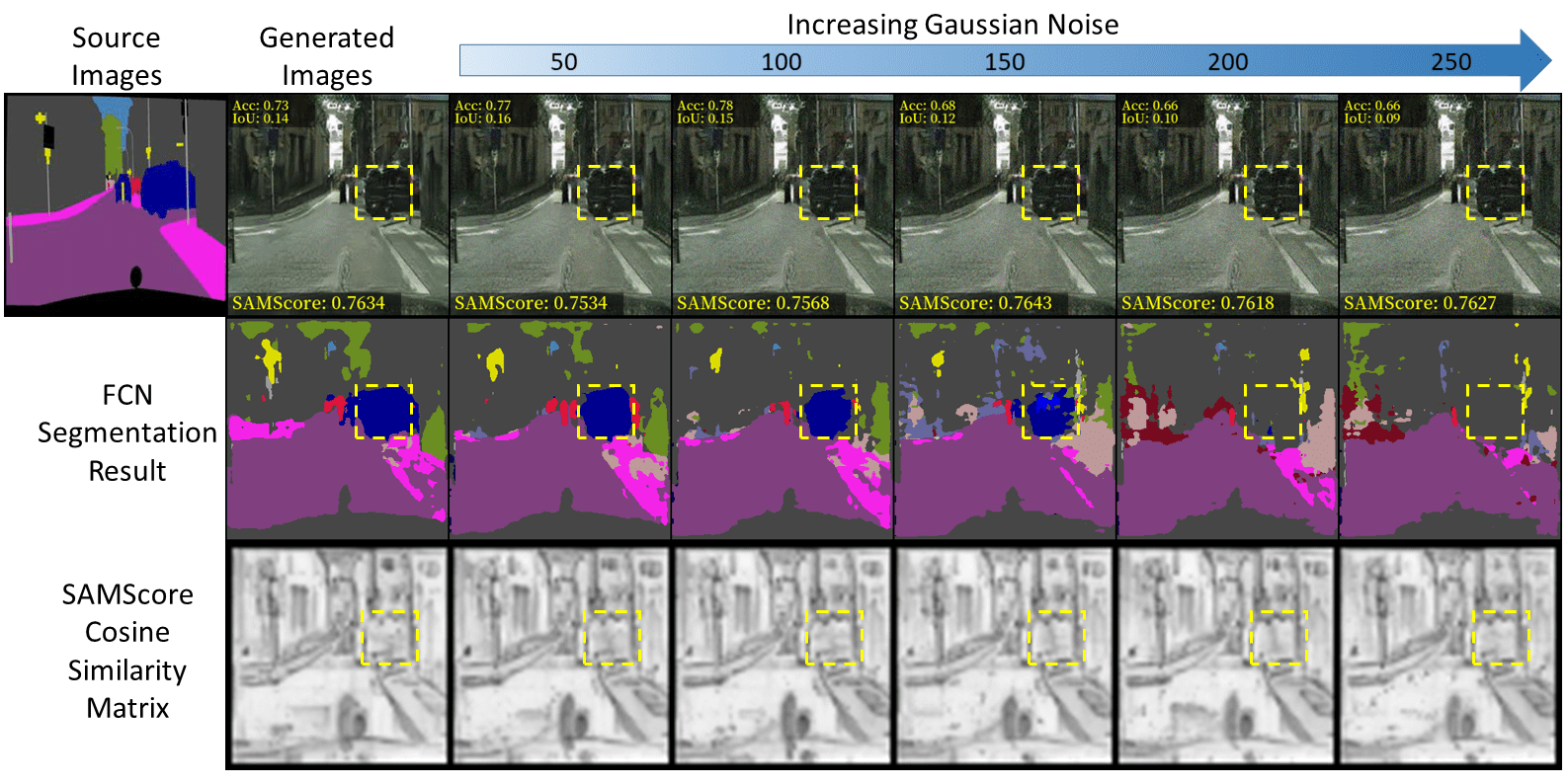}
  \caption{FCNScore and SAMScore: Gaussian noises were applied to the generated images of the cityscapes (label to photo) task, and then segmentation was performed by a FCN that had been trained on `true' target images. The content shape structural similarity of the generated images was then evaluated by applying accuracy and IoU metrics. Column 2 shows results without added distortion. Row 2 shows the segmentation results of FCN, and row 3 shows the cosine similarity matrix of SAMScore.}
  \label{fig:appendix_fcn2}
\end{figure*}

\begin{table*}[ht]
\centering
\renewcommand\arraystretch{1.1}
\caption{FCNScore (ACC and IoU) and SAMScore of translated images with different degrees of piecewise affine deformation. }
\begin{tabular}{c|cccccc}\hline
Degree of Deformation    & 0      & 0.01   & 0.02   & 0.03   & 0.04   & 0.05  \\\hline
Per-pixel ACC  & 0.5055 & 0.5055 & 0.5004 & 0.4973 & 0.4894 & 0.4828 \\
Class IoU  & 0.1248 & 0.1222 & 0.1175 & 0.1122 & 0.1049 & 0.0993 \\
SAMScore  & 0.7624 & 0.7498 & 0.7361 & 0.7247 & 0.7174 & 0.7091 \\ \hline
\end{tabular}
\label{mean_fcn_affine}
\end{table*}

\begin{table*}[ht]
\centering
\small
\renewcommand\arraystretch{1.1}
\caption{FCNScore (ACC and IoU) and SAMScore of translated images with different degrees of Gaussian noise corruption. }
\begin{tabular}{c|cccccc}\hline
Degree of Gaussian Noise & 0      & 50  & 100   & 150   & 200   & 250   \\\hline
Per-pixel ACC    & 0.5055 & 0.5013 & 0.4893 & 0.4714 & 0.4537 & 0.4402 \\
Class IoU     & 0.1248 & 0.1184 & 0.1087 & 0.0990 & 0.0896 & 0.0813 \\
SAMScore  & 0.7624 & 0.7476 & 0.7504 & 0.7530 & 0.7558 & 0.7574 \\\hline
\end{tabular}
\label{mean_fcn_gaussian}
\end{table*}

\begin{table}[ht]
\centering
\small
\renewcommand\arraystretch{1.1}
\caption{FCNScore (ACC and IoU) on target images and translated images. FCN is trained on `true' target domain images, and the cause for the large differences in FCNScore between the target image and the translated image is the presence of the domain gap (between the `true' target image domain and the translated image domain), which affects the segmentation accuracy on the translated images.}
\begin{tabular}{c|c|cccccc}\hline
  & Target Image & Translated Image \\ \hline 
Per-pixel ACC     & 0.8433 & 0.5055  \\
Class IoU   & 0.2711 & 0.1248  \\ \hline
\end{tabular}
\label{fcn_domain_gap}
\end{table}

\begin{figure*}[ht]
  \centering
  \includegraphics[width=.85\textwidth]{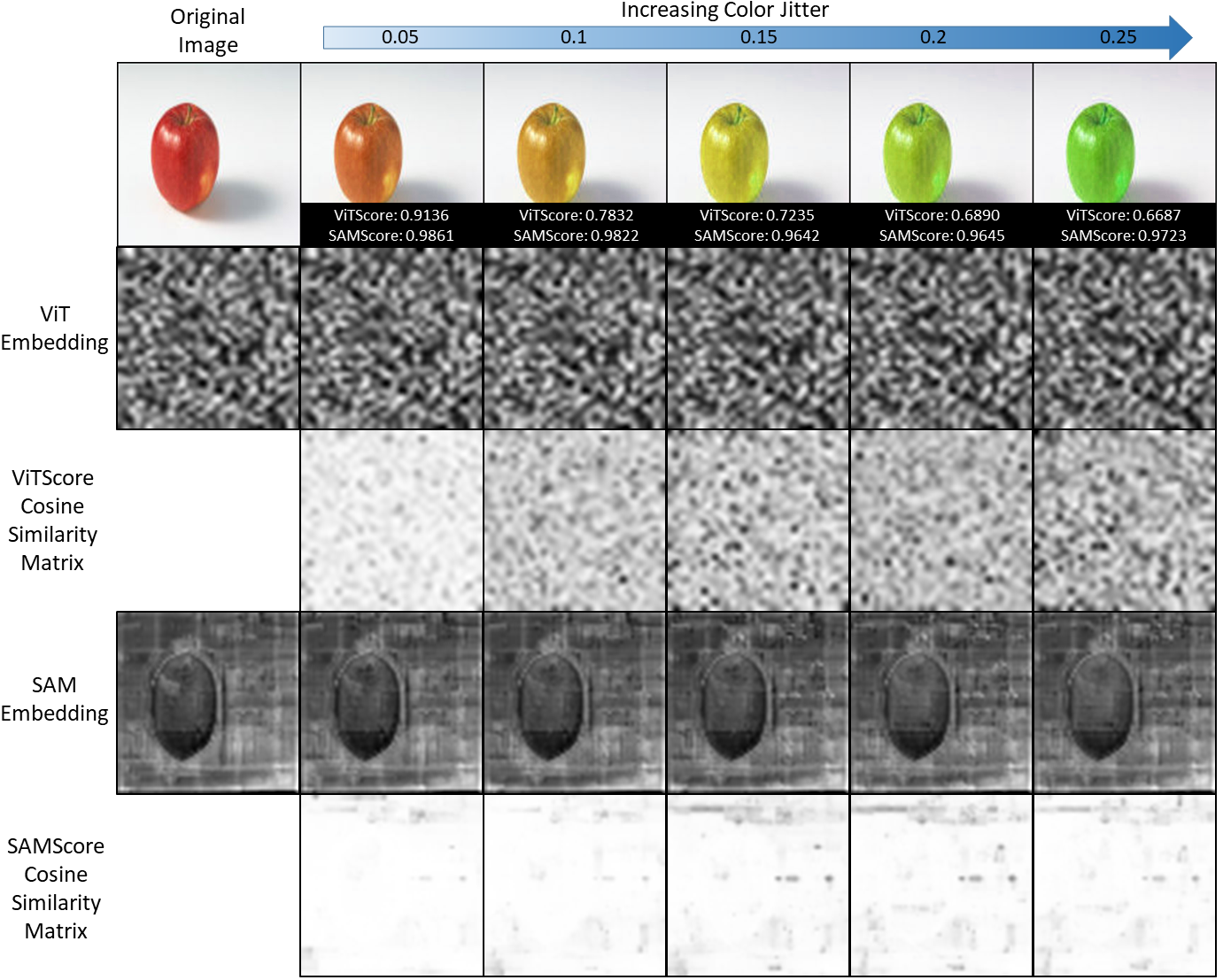}
  \caption{ViTScore and SAMScore: Color jitter was applied to the images of the apples, shown in row 1. Rows 2 and 4 show the embeddings of ViTScore and SAMScore with different degrees of color jitter. Rows 3 and 5 show the cosine similarity matrices with different degrees of color jitter.}
  \label{fig:vit_score}
\end{figure*}

\begin{table}[ht]
\caption{ViTScore and SAMScore of the apple images under different degrees of color jitter.}
\begin{tabular}{c|ccccc} \hline 
  Degree of Color Jitter       & 0.05   & 0.1    & 0.15   & 0.2    & 0.25   \\ \hline 
ViTScore & 0.9107 & 0.8015 & 0.7310 & 0.7007 & 0.6908 \\
SAMScore & 0.9944 & 0.9881 & 0.9831 & 0.9801 & 0.9803 \\ \hline 
\end{tabular}
\label{tab:vit_score}
\end{table}

\subsection{Comparison between SAMScore and FCNScore}

In FCNScore, the segmentation per-pixel accuracy (ACC) and the mean class Intersection-Over-Union (IoU) are two measures of the content shape structural similarity, using a pre-trained content segmentation network. To compare FCNScore against SAMScore, we obtained the segmentation ACC and IoU of the generated images using the FCN segmentation network (DeepLabV3Plus) on the cityscapes dataset, and compared the correlation coefficients of FCNScore and SAMScore against the amount of piecewise affine deformations and the degree of Gaussian noises. The results in Table \ref{corr_FCNScore} and Fig. \ref{fig:FCNScore} show that SAMScore outperformed FCNScore in both accuracy and robustness. The inferior performance of FCNScore may be attributed to the following two factors. First, there is a domain gap between the training data and the testing data. The training of the segmentation network used real images from the `true' target domain, but the measurement of the content structure was performed on the generated images, which do not have the exact same distribution as the target images, resulting in large segmentation errors that affect ACC and IoU. Secondly, the granularity of the structure that FCNScore can evaluate is limited by the granularity of the segmentation labels. For example, the segmentation labels in the cityscapes dataset are coarse and limited to large-scale objects. For instance, a car is segmented as a whole but its detailed sub-structures including windows and lights are not segmented. Using such coarse segmentation leads to the loss of critical information when comprehensively evaluating structural preservation and image similarity. Also, many image translation tasks do not have associated content segmentation labels, which limits the applicability of FCNScore.

In Tables \ref{mean_fcn_affine} and \ref{mean_fcn_gaussian}, we list the FCNScore (ACC and IoU) and SAMScore values for different degrees of piecewise affine distortions and Gaussian noise corruptions on the cityscapes (label to photo) task. It can be seen from Table \ref{fcn_domain_gap} that there is a large difference between the FCNScore on the translated image and on the target image. Since the FCN is trained using the `true' target domain image, the domain gap between the target images and the translated images renders it less efficient to extract the segmentations from the translated images for faithfulness evaluation.
In the cosine similarity matrix of SAMScore, brighter positions represent higher similarity, and darker positions represent lower similarity. Looking at the road safety posts in the dotted box in Fig. \ref{fig:appendix_fcn1}, the FCN cannot correctly segment them, resulting in inaccurate evaluations. In Fig. \ref{fig:appendix_fcn2}, the segmentation accuracy of FCN keeps decreasing with increasing Gaussian noise, while SAMScore shows robustness.
The sensitivity and robustness of SAMScore to such distortions and its ability to outperform traditional FCN-based measures suggest that it has the potential to be a valuable tool for evaluating and developing more sophisticated image translation models.

\begin{table*}[h]
\centering
\renewcommand\arraystretch{1.1}
\caption{Different similarity metrics on the medical image translation tasks (CT to MR and MR to CT). The arrows indicate the direction toward better performance. Best value is shown in bold. }
\begin{tabular}{cccccccc}
\hline 
\multicolumn{8}{c}{Head MR to CT} \\ \hline     
Method           &FID     $\downarrow$   & L2   $\downarrow$    & PSNR  $\uparrow$    & SSIM  $\uparrow$    & LPIPS  $\downarrow$   & SAMScore $\uparrow$ & L2 (GT) $\downarrow$ \\\hline
CycleGAN~\cite{zhu2017unpaired}  &138.99     & 113.79 & 12.91 & \textbf{0.5786} & 0.2294 & \textbf{0.8861} &\textbf{76.39}  \\
CycleDiffusion~\cite{wu2022unifying} &\textbf{85.42} & \textbf{89.06} & 15.13 & 0.5545 & \textbf{0.2290} & 0.8512 &116.81  \\
SDEdit~\cite{meng2021sdedit}       &122.53 & 75.64 & \textbf{16.59} & 0.5138 & 0.24833  & 0.8381 & 107.56   \\
EGSDE~\cite{zhao2022egsde}       &118.48  & 127.17 & 11.07 & 0.4033 & 0.4172 & 0.7780 &140.03  \\ \hline \hline
\multicolumn{8}{c}{Head CT to MR} \\   
\hline
Method           &FID     $\downarrow$   & L2   $\downarrow$    & PSNR  $\uparrow$    & SSIM  $\uparrow$    & LPIPS  $\downarrow$   & SAMScore $\uparrow$ & L2 (GT) $\downarrow$ \\\hline
CycleGAN~\cite{zhu2017unpaired}   &138.29    & 108.79 & 12.63 & \textbf{0.5285} & \textbf{0.1855}  & \textbf{0.9115} &\textbf{89.86} \\
CycleDiffusion~\cite{wu2022unifying} &\textbf{117.76}  & 77.69 & 17.18 & 0.3908 & 0.1995 & 0.8904 &111.22 \\
SDEdit~\cite{meng2021sdedit}      &126.33    & \textbf{60.66} & \textbf{19.27} & 0.3505 & 0.2228 & 0.8734& 112.62 \\
EGSDE~\cite{zhao2022egsde}      &123.30   & 134.34 & 11.13 & 0.3744 & 0.4047 & 0.8101 &133.36 \\ \hline
\end{tabular}
\label{head_mr2ct}
\end{table*}

\begin{figure*}[h]
  \centering
  \includegraphics[width=.85\textwidth]{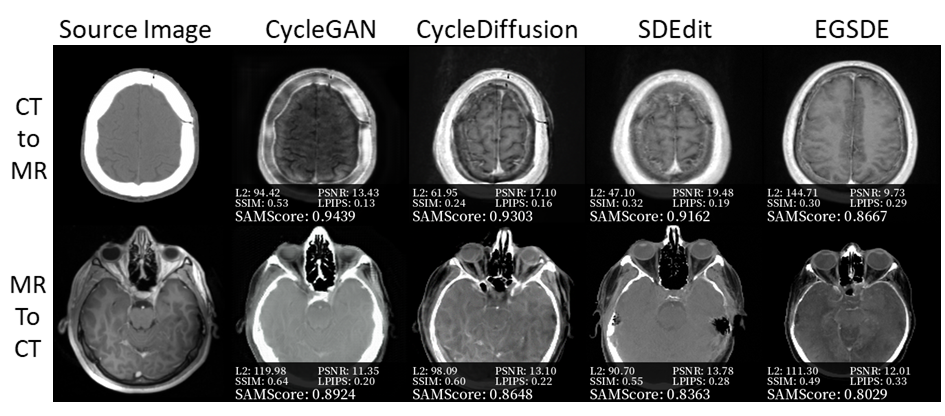}
  \caption{Evaluation of compared image similarity metrics on image translation tasks conducted by multiple state-of-the-art generative models, including CycleGAN, CycleDiffusion, SDEdit, and EGSDE, applied on head CT and MR images.}
  \label{fig:head}
\end{figure*}

\subsection{Comparison between SAMScore and ViTScore}
Taking into account that the network structure of SAM’s encoder is based on the Vision Transformer (ViT), we also visualized the embeddings of a ViT model that was pre-trained on ImageNet-21k for the classification task, and compared those with the SAM embeddings. We named the cosine similarity score calculated on the ViT embeddings as ViTScore. As shown in Fig. \ref{fig:vit_score}, ViTScore's embeddings do not contain clear content shapes. This suggests that while the ViT model uses a similar network structure, its embeddings may not be as suited as SAMScore's embeddings for evaluating content structural similarities, since SAM is specifically trained for segmentation to capture the content shapes. The power of the SAM model in capturing the content shapes is mostly due to the large dataset used to train the specific segmentation task, rather than the network design.

In addition, we used the apple-to-orange dataset to evaluate the robustness of ViTScore and SAMScore to increasing color jitter, which was realized through the Albumentations data augmentation library~\cite{buslaev2020albumentations}. We did not use the transferred orange image (from the apple image) for this color jitter study, since the apple and the orange are of very different colors and it is difficult to create scenarios of increasing color jitters relative to the original apple image by using the transferred orange image. Thus, we applied the color jitter directly on the original apple image to create increasing degrees of variations. It can be seen from Fig. \ref{fig:vit_score} and Table \ref{tab:vit_score} that with increasing degrees of color dithering, the cosine similarity matrix values of the ViT model’s embeddings and the corresponding ViTScore dropped significantly, while the corresponding results for the SAMScore were barely affected. The results indicate that SAM's embeddings are robust to color variations, and are more focused on the content structure of the image rather than its color or texture compositions.

\begin{table*}[h]
\centering
\renewcommand\arraystretch{1.1}
\caption{Performance of images generated by Swin Transformer with different levels of Gaussian noise added to the input source image. The arrows indicate the direction toward better performance. Changes relative to the previous noise level are shown in the adjacent columns.}
\begin{tabular}{cccccccccccc}
\hline  
\multirow{2}{*}{Noise Level}  & \multicolumn{2}{c}{L2 $\downarrow$}    & \multicolumn{2}{c}{PSNR $\uparrow$}    & \multicolumn{2}{c}{SSIM $\uparrow$}    & \multicolumn{2}{c}{LPIPS $\downarrow$}   & \multicolumn{2}{c}{SAMScore $\uparrow$} \\\cline{2-11}
             & Value  & Change  & Value  & Change  & Value  & Change  & Value  & Change  & Value  & Change  \\\hline
0     & 72.0932  & -        & 16.0783  & -        & 0.7733 & -         & 0.2034 & -         & 0.9245  & -         \\
50  & 71.7789  & -0.3143  & 16.1252  & +0.0469  & 0.7767 & +0.0034  & 0.2033 & -0.0001  & 0.9151  & -0.0094  \\
100  & 71.6133  & -0.1655  & 16.1515  & +0.0263  & 0.7784 & +0.0017  & 0.2045 & +0.0012  & 0.9112  & -0.0039  \\
150  & 71.5043  & -0.1090  & 16.1681  & +0.0166  & 0.7794 & +0.0010  & 0.2060 & +0.0014  & 0.9082  & -0.0030  \\
200  & 71.4146  & -0.0898  & 16.1829  & +0.0149  & 0.7800 & +0.0006  & 0.2073 & +0.0013  & 0.9057  & -0.0025  \\
250   & 71.3520  & -0.0626  & 16.1925  & +0.0096  & 0.7805 & +0.0005  & 0.2084 & +0.0011  & 0.9040  & -0.0018  \\\hline 
\end{tabular}
\label{tab:noise_swin_performance}
\end{table*}

\begin{figure*}[h]
  \centering
  \includegraphics[width=.99\textwidth]{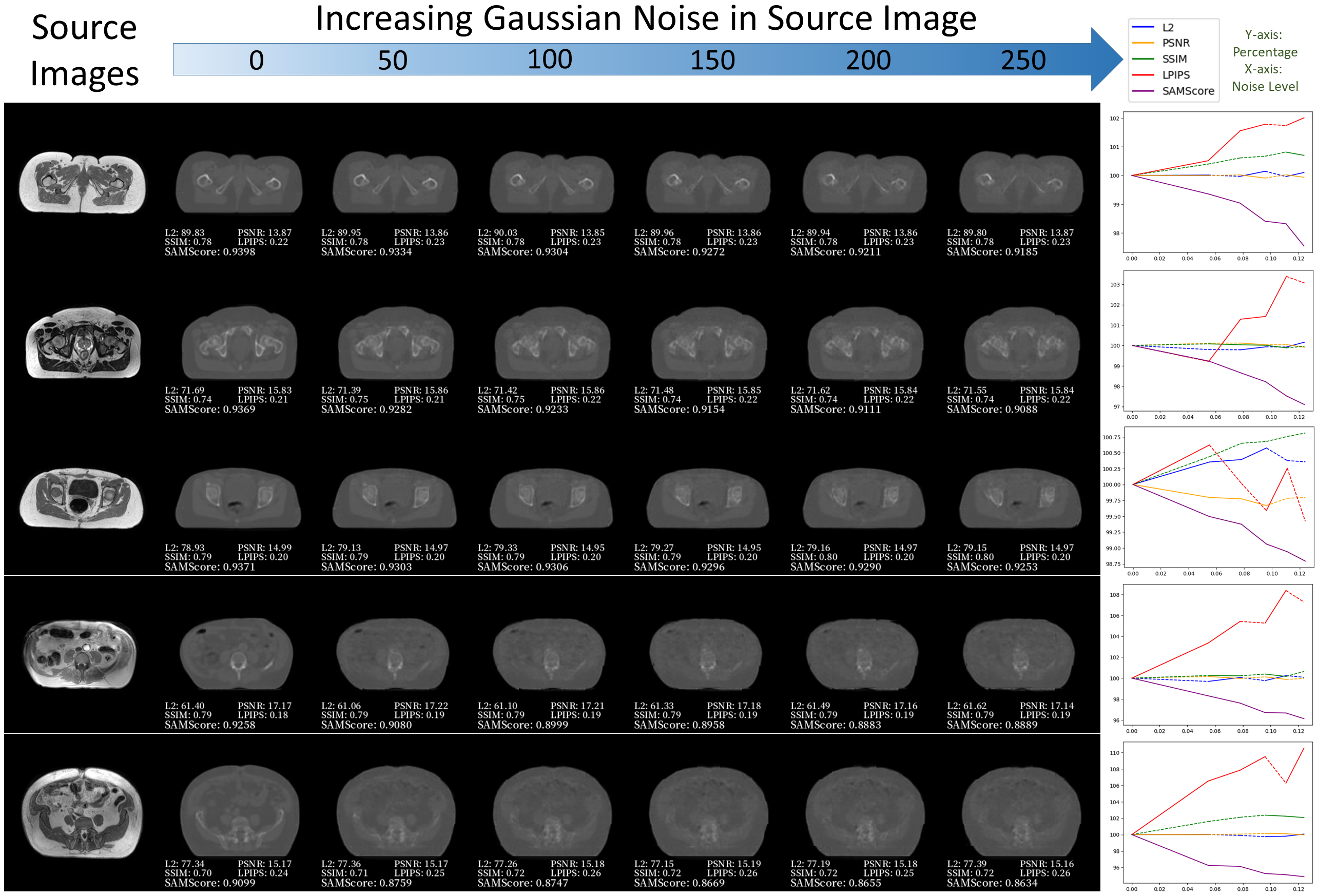}
  \caption{Visualization of images generated by Swin Transformer with different levels of Gaussian noise added to the input source image, along with their corresponding similarity scores. Given the differing scales of the metrics, we converted each metric to a percentage change relative to the initial value (results without any perturbation) before generating the line plots. We used dashed lines to represent trends that deviate from the ideal direction of change.}
  \label{fig:noise_swin_performance}
\end{figure*}

\subsection{Performance on Different Translation Models}
We also applied each of the similarity metrics to evaluate different image translation methods, for the purpose of assessing whether a metric is able to consistently measure the performance of image translation models. A medical image translation task (Head CT to MR and MR to CT translations) was used for this test. CycleGAN, the most widely used image translation benchmark, was evaluated. In comparison, the three diffusion-based methods, CycleDiffusion, SDEdit, and EGSDE, were also evaluated. From the visual comparison in Fig. \ref{fig:head} and the L2 results with ground truth (GT) in Table \ref{head_mr2ct}, the CycleGAN model appeared to best preserve the source structure, as it uses a consistency loss function to promote the retention of content information. The results in Table \ref{head_mr2ct} show that, based on the SAMScore, CycleGAN performed the best on both the CT to MR and the MR to CT translation tasks, which is consistent with visual evaluations and L2 (GT). The other metrics, especially L2 and PSNR, failed to provide consistent measurements with visual evaluation and L2 (GT). CT and MR present very different pixel values on the same underlying structures, rendering the pixel-based L2 and PSNR measures ineffective and highly error-prone. Among all faithfulness metrics, only the SAMScore ranking for each model is consistent with L2 (GT). The Frechet Inception Distance (FID)~\cite{heusel2017gans}, a widely used metric for assessing realism, is computed by comparing a generated image dataset with a target domain dataset. Although the translation models based on diffusion produced less faithful results than CycleGAN, they performed better in terms of realism. This suggests that the joint optimization of FID and SAMScore could facilitate the emergence of translation models that perform better in terms of both faithfulness and realism.

\subsection{Performance Under Different Image Qualities}
To evaluate whether SAMScore can effectively assess the quality of generated images, we performed experiments to systematically study the consistency of evaluation metrics with varying visual image qualities. Specifically, we used the Swin Transformer backbone trained in a paired fashion on the SynthRAD pelvic MR-to-CT dataset to establish a powerful translation model. During inference, Gaussian noise was artificially added to the input images to generate different outputs with varying quality. These outputs ranged from high-quality to low-quality, providing a `sliding-bar' baseline to evaluate the sensitivity of varying evaluation metrics. For each noise level, the generated results were compared with the input source image, and multiple metrics including L2, PSNR, SSIM, LPIPS, and SAMScore were calculated to analyze their trends as image quality declined. The results, as shown in Table \ref{tab:noise_swin_performance}, indicate that SAMScore accurately reflects changes in image quality, consistently showing a downward trend as noise levels increase, aligning with perceptual changes in image quality. In contrast, other common metrics such as L2, SSIM, PSNR, and LPIPS, often failed to adequately reflect the degradation in image quality and showed inconsistent trends. For example, while higher SSIM and PSNR values indicate better performance, both metrics paradoxically increased as image quality declined, contradicting the actual perceptual changes. In Fig. \ref{fig:noise_swin_performance}, as noise levels increase, the skeletal structures in the generated images become more blurred, and the boundaries of soft tissues gradually disappear, indicating structural content changes in the images. Our SAMScore effectively captures this decline in image quality, whereas other metrics struggle to represent and quantify such changes. 
The experiments show that SAMScore demonstrates a good alignment with human perception under varying image quality conditions. This highlights SAMScore's capability of evaluating translated images of different quality levels, offering a reliable and consistent standard for image translation tasks.

\section{Limitations and Discussion}
Although SAMScore has demonstrated strong evaluation capabilities in various image translation tasks, we acknowledge its limitations in certain specific application scenarios. For instance, in bone MR-to-CT translation tasks, dense bones appear dark in MR due to the absence of water, whereas in CT, they are less distinguishable from surrounding bone tissues. This discrepancy can even confuse experienced radiologists. Additionally, the prostate exhibits different sizes in MR and CT due to differences in imaging mechanisms. MR is more sensitive to soft tissue contrast, providing a clearer representation of prostate details, while CT is less capable of enhancing prostate boundaries. Consequently, the prostate often appears larger in MR compared to CT, which may lead to inaccuracies in SAMScore evaluations. These specific cases highlight the importance of carefully assessing whether the assumptions underlying SAMScore (invariant structures before/after translation) are appropriate for the intended application. Users should perform a thorough feasibility analysis before employing SAMScore in their particular scenarios.

\section{Conclusion and Future Work}
We presented SAMScore, a generic content structural similarity metric for evaluating image translation tasks. Our experimental results demonstrate that SAMScore substantially outperforms traditional metrics in terms of accuracy and robustness in evaluating the content structural similarity. With the ability to better distinguish and evaluate translation models in terms of the retention of content structures, we believe that SAMScore has the potential to significantly contribute to the development of more effective techniques in the vibrant research area of image-to-image translation. We envision a variety of interesting avenues for future work, including the assessment of image structural similarity for specified regions of interest, and the development of SAMScore-guided image translation models.

\bibliographystyle{IEEEtran}
\bibliography{paper1.bib}

\begin{thebibliography}{10}
\providecommand{\url}[1]{#1}
\csname url@samestyle\endcsname
\providecommand{\newblock}{\relax}
\providecommand{\bibinfo}[2]{#2}
\providecommand{\BIBentrySTDinterwordspacing}{\spaceskip=0pt\relax}
\providecommand{\BIBentryALTinterwordstretchfactor}{4}
\providecommand{\BIBentryALTinterwordspacing}{\spaceskip=\fontdimen2\font plus
\BIBentryALTinterwordstretchfactor\fontdimen3\font minus
  \fontdimen4\font\relax}
\providecommand{\BIBforeignlanguage}[2]{{%
\expandafter\ifx\csname l@#1\endcsname\relax
\typeout{** WARNING: IEEEtran.bst: No hyphenation pattern has been}%
\typeout{** loaded for the language `#1'. Using the pattern for}%
\typeout{** the default language instead.}%
\else
\language=\csname l@#1\endcsname
\fi
#2}}
\providecommand{\BIBdecl}{\relax}
\BIBdecl

\bibitem{isola2017image}
P.~Isola, J.-Y. Zhu, T.~Zhou, and A.~A. Efros, ``Image-to-image translation
  with conditional adversarial networks,'' in \emph{Proceedings of the IEEE
  conference on computer vision and pattern recognition}, 2017, pp. 1125--1134.

\bibitem{liu2017unsupervised}
M.-Y. Liu, T.~Breuel, and J.~Kautz, ``Unsupervised image-to-image translation
  networks,'' \emph{Advances in neural information processing systems},
  vol.~30, 2017.

\bibitem{zhu2017toward}
J.-Y. Zhu, R.~Zhang, D.~Pathak, T.~Darrell, A.~A. Efros, O.~Wang, and
  E.~Shechtman, ``Toward multimodal image-to-image translation,''
  \emph{Advances in neural information processing systems}, vol.~30, 2017.

\bibitem{choi2018stargan}
Y.~Choi, M.~Choi, M.~Kim, J.-W. Ha, S.~Kim, and J.~Choo, ``Stargan: Unified
  generative adversarial networks for multi-domain image-to-image
  translation,'' in \emph{Proceedings of the IEEE conference on computer vision
  and pattern recognition}, 2018, pp. 8789--8797.

\bibitem{richardson2021encoding}
E.~Richardson, Y.~Alaluf, O.~Patashnik, Y.~Nitzan, Y.~Azar, S.~Shapiro, and
  D.~Cohen-Or, ``Encoding in style: a stylegan encoder for image-to-image
  translation,'' in \emph{Proceedings of the IEEE/CVF conference on computer
  vision and pattern recognition}, 2021, pp. 2287--2296.

\bibitem{armanious2020medgan}
K.~Armanious, C.~Jiang, M.~Fischer, T.~K{\"u}stner, T.~Hepp, K.~Nikolaou,
  S.~Gatidis, and B.~Yang, ``Medgan: Medical image translation using gans,''
  \emph{Computerized medical imaging and graphics}, vol.~79, p. 101684, 2020.

\bibitem{tumanyan2023plug}
N.~Tumanyan, M.~Geyer, S.~Bagon, and T.~Dekel, ``Plug-and-play diffusion
  features for text-driven image-to-image translation,'' in \emph{Proceedings
  of the IEEE/CVF Conference on Computer Vision and Pattern Recognition}, 2023,
  pp. 1921--1930.

\bibitem{li2023bbdm}
B.~Li, K.~Xue, B.~Liu, and Y.-K. Lai, ``Bbdm: Image-to-image translation with
  brownian bridge diffusion models,'' in \emph{Proceedings of the IEEE/CVF
  conference on computer vision and pattern Recognition}, 2023, pp. 1952--1961.

\bibitem{lee2024conditional}
H.~Lee, M.~Kang, and B.~Han, ``Conditional score guidance for text-driven
  image-to-image translation,'' \emph{Advances in Neural Information Processing
  Systems}, vol.~36, 2024.

\bibitem{xu2024cyclenet}
S.~Xu, Z.~Ma, Y.~Huang, H.~Lee, and J.~Chai, ``Cyclenet: Rethinking cycle
  consistency in text-guided diffusion for image manipulation,'' \emph{Advances
  in Neural Information Processing Systems}, vol.~36, 2024.

\bibitem{pang2021image}
Y.~Pang, J.~Lin, T.~Qin, and Z.~Chen, ``Image-to-image translation: Methods and
  applications,'' \emph{IEEE Transactions on Multimedia}, vol.~24, pp.
  3859--3881, 2021.

\bibitem{kaji2019overview}
S.~Kaji and S.~Kida, ``Overview of image-to-image translation by use of deep
  neural networks: denoising, super-resolution, modality conversion, and
  reconstruction in medical imaging,'' \emph{Radiological physics and
  technology}, vol.~12, pp. 235--248, 2019.

\bibitem{alotaibi2020deep}
A.~Alotaibi, ``Deep generative adversarial networks for image-to-image
  translation: A review,'' \emph{Symmetry}, vol.~12, no.~10, p. 1705, 2020.

\bibitem{arar2020unsupervised}
M.~Arar, Y.~Ginger, D.~Danon, A.~H. Bermano, and D.~Cohen-Or, ``Unsupervised
  multi-modal image registration via geometry preserving image-to-image
  translation,'' in \emph{Proceedings of the IEEE/CVF conference on computer
  vision and pattern recognition}, 2020, pp. 13\,410--13\,419.

\bibitem{zhang2018unreasonable}
R.~Zhang, P.~Isola, A.~A. Efros, E.~Shechtman, and O.~Wang, ``The unreasonable
  effectiveness of deep features as a perceptual metric,'' in \emph{Proceedings
  of the IEEE conference on computer vision and pattern recognition}, 2018, pp.
  586--595.

\bibitem{borji2019pros}
A.~Borji, ``Pros and cons of gan evaluation measures,'' \emph{Computer Vision
  and Image Understanding}, vol. 179, pp. 41--65, 2019.

\bibitem{bashkirova2022evaluation}
D.~Bashkirova, B.~Usman, and K.~Saenko, ``Evaluation of correctness in
  unsupervised many-to-many image translation,'' in \emph{Proceedings of the
  IEEE/CVF Winter Conference on Applications of Computer Vision}, 2022, pp.
  1776--1785.

\bibitem{wang2004image}
Z.~Wang, A.~C. Bovik, H.~R. Sheikh, and E.~P. Simoncelli, ``Image quality
  assessment: from error visibility to structural similarity,'' \emph{IEEE
  transactions on image processing}, vol.~13, no.~4, pp. 600--612, 2004.

\bibitem{lahouhou2010selecting}
A.~Lahouhou, E.~Viennet, and A.~Beghdadi, ``Selecting low-level features for
  image quality assessment by statistical methods,'' \emph{Journal of computing
  and information technology}, vol.~18, no.~2, pp. 183--189, 2010.

\bibitem{zhu2017unpaired}
J.-Y. Zhu, T.~Park, P.~Isola, and A.~A. Efros, ``Unpaired image-to-image
  translation using cycle-consistent adversarial networks,'' in
  \emph{Proceedings of the IEEE international conference on computer vision},
  2017, pp. 2223--2232.

\bibitem{zhangbertscore}
T.~Zhang, V.~Kishore, F.~Wu, K.~Q. Weinberger, and Y.~Artzi, ``Bertscore:
  Evaluating text generation with bert,'' in \emph{International Conference on
  Learning Representations}.

\bibitem{hessel2021clipscore}
J.~Hessel, A.~Holtzman, M.~Forbes, R.~L. Bras, and Y.~Choi, ``Clipscore: A
  reference-free evaluation metric for image captioning,'' \emph{arXiv preprint
  arXiv:2104.08718}, 2021.

\bibitem{kirillov2023segment}
A.~Kirillov, E.~Mintun, N.~Ravi, H.~Mao, C.~Rolland, L.~Gustafson, T.~Xiao,
  S.~Whitehead, A.~C. Berg, W.-Y. Lo \emph{et~al.}, ``Segment anything,''
  \emph{arXiv preprint arXiv:2304.02643}, 2023.

\bibitem{MedSAM}
J.~Ma and B.~Wang, ``Segment anything in medical images,'' \emph{arXiv preprint
  arXiv:2304.12306}, 2023.

\bibitem{zhang2019gcgan}
Y.~Zhang, S.~Wang, B.~Chen, and J.~Cao, ``Gcgan: Generative adversarial nets
  with graph cnn for network-scale traffic prediction,'' in \emph{2019
  International Joint Conference on Neural Networks (IJCNN)}.\hskip 1em plus
  0.5em minus 0.4em\relax IEEE, 2019, pp. 1--8.

\bibitem{park2020contrastive}
T.~Park, A.~A. Efros, R.~Zhang, and J.-Y. Zhu, ``Contrastive learning for
  unpaired image-to-image translation,'' in \emph{Computer Vision--ECCV 2020:
  16th European Conference, Glasgow, UK, August 23--28, 2020, Proceedings, Part
  IX 16}.\hskip 1em plus 0.5em minus 0.4em\relax Springer, 2020, pp. 319--345.

\bibitem{huang2018multimodal}
X.~Huang, M.-Y. Liu, S.~Belongie, and J.~Kautz, ``Multimodal unsupervised
  image-to-image translation,'' in \emph{Proceedings of the European conference
  on computer vision (ECCV)}, 2018, pp. 172--189.

\bibitem{choi2020stargan}
Y.~Choi, Y.~Uh, J.~Yoo, and J.-W. Ha, ``Stargan v2: Diverse image synthesis for
  multiple domains,'' in \emph{Proceedings of the IEEE/CVF conference on
  computer vision and pattern recognition}, 2020, pp. 8188--8197.

\bibitem{paolacci2010running}
G.~Paolacci, J.~Chandler, and P.~G. Ipeirotis, ``Running experiments on amazon
  mechanical turk,'' \emph{Judgment and Decision making}, vol.~5, no.~5, pp.
  411--419, 2010.

\bibitem{li2022vqbb}
B.~Li, K.~Xue, B.~Liu, and Y.-K. Lai, ``Vqbb: Image-to-image translation with
  vector quantized brownian bridge,'' \emph{arXiv preprint arXiv:2205.07680},
  2022.

\bibitem{meng2021sdedit}
C.~Meng, Y.~He, Y.~Song, J.~Song, J.~Wu, J.-Y. Zhu, and S.~Ermon, ``Sdedit:
  Guided image synthesis and editing with stochastic differential equations,''
  in \emph{International Conference on Learning Representations}, 2021.

\bibitem{wu2022unifying}
C.~H. Wu and F.~De~la Torre, ``Unifying diffusion models' latent space, with
  applications to cyclediffusion and guidance,'' \emph{arXiv preprint
  arXiv:2210.05559}, 2022.

\bibitem{zhao2022egsde}
M.~Zhao, F.~Bao, C.~Li, and J.~Zhu, ``Egsde: Unpaired image-to-image
  translation via energy-guided stochastic differential equations,''
  \emph{arXiv preprint arXiv:2207.06635}, 2022.

\bibitem{shusharina2021segmentation}
N.~Shusharina, M.~P. Heinrich, and R.~Huang, \emph{Segmentation,
  Classification, and Registration of Multi-modality Medical Imaging Data:
  MICCAI 2020 Challenges, ABCs 2020, L2R 2020, TN-SCUI 2020, Held in
  Conjunction with MICCAI 2020, Lima, Peru, October 4--8, 2020,
  Proceedings}.\hskip 1em plus 0.5em minus 0.4em\relax Springer Nature, 2021,
  vol. 12587.

\bibitem{cordts2016cityscapes}
M.~Cordts, M.~Omran, S.~Ramos, T.~Rehfeld, M.~Enzweiler, R.~Benenson,
  U.~Franke, S.~Roth, and B.~Schiele, ``The cityscapes dataset for semantic
  urban scene understanding,'' in \emph{Proceedings of the IEEE conference on
  computer vision and pattern recognition}, 2016, pp. 3213--3223.

\bibitem{tylevcek2013spatial}
R.~Tyle{\v{c}}ek and R.~{\v{S}}{\'a}ra, ``Spatial pattern templates for
  recognition of objects with regular structure,'' in \emph{Pattern
  Recognition: 35th German Conference, GCPR 2013, Saarbr{\"u}cken, Germany,
  September 3-6, 2013. Proceedings 35}.\hskip 1em plus 0.5em minus 0.4em\relax
  Springer, 2013, pp. 364--374.

\bibitem{deng2009imagenet}
J.~Deng, W.~Dong, R.~Socher, L.-J. Li, K.~Li, and L.~Fei-Fei, ``Imagenet: A
  large-scale hierarchical image database,'' in \emph{2009 IEEE conference on
  computer vision and pattern recognition}.\hskip 1em plus 0.5em minus
  0.4em\relax Ieee, 2009, pp. 248--255.

\bibitem{thummerer2023synthrad2023}
A.~Thummerer, E.~van~der Bijl, A.~Galapon~Jr, J.~J. Verhoeff, J.~A. Langendijk,
  S.~Both, C.~N.~A. van~den Berg, and M.~Maspero, ``Synthrad2023 grand
  challenge dataset: Generating synthetic ct for radiotherapy,'' \emph{Medical
  physics}, vol.~50, no.~7, pp. 4664--4674, 2023.

\bibitem{pitiot2003piecewise}
A.~Pitiot, G.~Malandain, E.~Bardinet, and P.~M. Thompson, ``Piecewise affine
  registration of biological images,'' in \emph{Biomedical Image Registration:
  Second InternationalWorkshop, WBIR 2003, Philadelphia, PA, USA, June 23-24,
  2003. Revised Papers 2}.\hskip 1em plus 0.5em minus 0.4em\relax Springer,
  2003, pp. 91--101.

\bibitem{qin2010multivariate}
A.~K. Qin and D.~A. Clausi, ``Multivariate image segmentation using semantic
  region growing with adaptive edge penalty,'' \emph{IEEE transactions on image
  processing}, vol.~19, no.~8, pp. 2157--2170, 2010.

\bibitem{IDDPM}
A.~Q. Nichol and P.~Dhariwal, ``Improved denoising diffusion probabilistic
  models,'' in \emph{International Conference on Machine Learning}, 2021, pp.
  8162--8171.

\bibitem{kingma2014adam}
D.~P. Kingma and J.~Ba, ``Adam: A method for stochastic optimization,'' in
  \emph{International Conference on Learning Representations}, 2014.

\bibitem{buslaev2020albumentations}
A.~Buslaev, V.~I. Iglovikov, E.~Khvedchenya, A.~Parinov, M.~Druzhinin, and
  A.~A. Kalinin, ``Albumentations: fast and flexible image augmentations,''
  \emph{Information}, vol.~11, no.~2, p. 125, 2020.

\bibitem{liu2021swin}
Z.~Liu, Y.~Lin, Y.~Cao, H.~Hu, Y.~Wei, Z.~Zhang, S.~Lin, and B.~Guo, ``Swin
  transformer: Hierarchical vision transformer using shifted windows,'' in
  \emph{Proceedings of the IEEE/CVF international conference on computer
  vision}, 2021, pp. 10\,012--10\,022.

\bibitem{heusel2017gans}
M.~Heusel, H.~Ramsauer, T.~Unterthiner, B.~Nessler, and S.~Hochreiter, ``Gans
  trained by a two time-scale update rule converge to a local nash
  equilibrium,'' \emph{Advances in neural information processing systems},
  vol.~30, 2017.

\end{thebibliography}

\end{document}